\documentclass{article}

\usepackage{microtype}
\usepackage{graphicx}
\usepackage{subfigure}
\usepackage{booktabs} 
\usepackage{xcolor}
\usepackage{float}
\usepackage{makecell}
\usepackage{diagbox}
\usepackage{placeins}
\usepackage{afterpage}
\usepackage{enumitem}
\usepackage[linesnumbered]{algorithm2e}

\usepackage{hyperref}

\usepackage[accepted]{icml2024}

\usepackage{amsmath}
\usepackage{amssymb}
\usepackage{mathtools}
\usepackage{amsthm}

\usepackage[capitalize,noabbrev]{cleveref}

\theoremstyle{plain}

\theoremstyle{definition}

\theoremstyle{remark}

\usepackage[textsize=tiny]{todonotes}

\icmltitlerunning{SQBC}

\begin{document}

\twocolumn[
	\icmltitle{SQBC: Active Learning using LLM-Generated Synthetic Data \\ 
	for Stance Detection in Online Political Discussions}




	\begin{icmlauthorlist}
		\icmlauthor{Stefan Sylvius Wagner}{hhu}
		\icmlauthor{Maike Behrendt}{hhu}
		\icmlauthor{Marc Ziegele}{ssc}
		\icmlauthor{Stefan Harmeling}{tu}
		
	\end{icmlauthorlist}

	\icmlaffiliation{hhu}{Department of Computer Science, Heinrich Heine University Düsseldorf, Germany}
	\icmlaffiliation{ssc}{Department of Social Sciences, Heinrich Heine University Düsseldorf, Germany}
	\icmlaffiliation{tu}{Department of Computer Science, Technical University of Dortmund, Germany}

	\icmlcorrespondingauthor{Stefan Sylvius Wagner}{stefan.wagner@hhu.de}

	\icmlkeywords{Machine Learning, ICML}

	\vskip 0.3in
]



\printAffiliationsAndNotice{}  

\begin{abstract}
  Stance detection is an important task for many applications that analyse or support online political discussions.  
  Common approaches include fine-tuning transformer based models.  However, these models require a large amount of labelled data, which might not be available.
  In this work, we present two different ways to leverage LLM-generated synthetic data to train and improve stance detection agents for online political discussions:
  first, we show that augmenting a small fine-tuning dataset with synthetic data can improve the performance of the stance detection model. 
  Second, we propose a new active learning method called SQBC based on the "Query-by-Comittee" approach.  The key idea is to use LLM-generated synthetic data as an oracle to identify the most informative unlabelled samples, that are selected for manual labelling. 
  Comprehensive experiments show that both ideas can improve the stance detection performance.  Curiously, we observed that fine-tuning on actively selected samples can exceed the performance of using the full dataset.
\end{abstract}

\section{Introduction}
With the recent release of powerful generative Large Language Models (LLMs) such as ChatGPT\footnote{\url{https://chat.openai.com/}} and
Llama \citep{touvron2023llama}, research efforts have gone into investigating how synthetic data generated 
from LLMs can be leveraged for downstream tasks. It is well known that Natural Language Processing (NLP) tasks based on single-label or 
multi-label classification such as stance and sentiment detection
benefit immensely from large amounts of labelled data~\citep{mehrafarin-etal-2022-importance}. However, in many cases, labelled data is scarce, especially in
the realm of online political discussions, where gathering labelled data may not be allowed for privacy reasons.
Other challenges in stance detection for online political discussions include the fact that the stance of a statement is intrinsically
linked to a topic, since there is usually an additional question (sometimes also called issue or target) linked to each comment. This is problematic, since the samples may not be varied or pertain to only one class, hence usually yielding an unbalanced dataset.
Especially in stance detection for online political discussions, there are a potentially infinite number of specific questions
open for debate. For instance,  performance drops severely for questions in the training dataset, when the existing data does not reflect the question at hand \citep{vamvas2020xstance}.
With this in mind, we see the possibility of leveraging synthetic data generated from LLMs to improve stance detection models for  online political discussions.
We propose two ways in which synthetic data can be used to improve stance detection models for online political discussions:
\begin{enumerate}[leftmargin=*]
  \item \textbf{Fine-tuning with synthetic data.} One problem of stance detection datasets for online political discussions is the lack of (balanced) data for specific questions (issues).
        We propose to simply augment the existing data with synthetic data to fine-tune a stance detection model.
        This alone is a powerful tool to improve the performance of a stance detection model by grounding it to a specific question.
  \item \textbf{Active learning with synthetic data.} Another challenge in stance detection for online political discussion is labelling new data. Labelling data 
   is expensive and time consuming and it is therefore desirable to reduce labelling effort. For this purpose, we also propose a new active learning method (called Synthetic Data-driven Query By Comittee (SQBC)) using LLM-generated synthetic data.
        We use the synthetic data as an oracle to infer the most interesting unlabelled data points by checking the similarity between the embeddings of the unlabelled data to the synthetic data. This set of unlabelled data points is smaller while providing
        the same performance as if labelling all data points.
\end{enumerate}
We adopt a testing scenario similar to \citet{romberg2022automated} where we try to reduce the number of data points that need to be labelled by a human.
We show that we can reduce the labelling effort substantially by performing active learning with synthetic data and then augmenting the labelled data with synthetic data.

\section{Background}

\subsection{Stance Detection for Online Political Discussions}
Stance detection is a sub-task of sentiment analysis \citep{10.1145/3603254} and opinion mining \citep{ALDAYEL2021102597} that aims to automatically detect the stance (e.g., \emph{favor}, \emph{against}, or \emph{neutral}) of the author of a contribution towards a discussed issue, or target. In online political discussions, stances can be defined as the \emph{side} that the author is on, e.g., \emph{for} or \emph{against} an increase in value added tax. Research has identified stance detection as an important yet challenging task; it is important because the automated detection of stances can, e.g., improve discussion summarisation \citep{CHOWANDA201711}, facilitate the detection of misinformation \citep{hardalov-etal-2022-survey}, and provide more comprehensive evaluations of opinion distributions in online political discussions and participation processes~\citep{10.1145/3603254}. Stance detection is also implemented as a feature in many recommender systems and discussion platforms \citep{10.1145/3369026}. Still, due to its dependency on context, stance detection is a highly challenging task. In online political discussions, for example, detecting the stance of a contribution towards a question requires a proper identification of both the question under discussion and the position of the contributor towards it. However, users regularly deviate from the original question and discuss multiple topics in the same thread \citep{10.1111/jcom.12123}, resulting in a high number of potential questions with often very few instances for training purposes. 

Fine-tuning transformer-based models \citep{NIPS2017_3f5ee243}  to solve stance detection is a common practice, but training these models requires a large amount of annotated data. In our work, we show how to improve stance detection for online political discussions with synthetic data. 

\subsection{Active Learning}
 The aim of \emph{active learning} is to minimize the effort of labelling data, while simultaneously maximizing the model's performance. This is achieved by letting the model choose the most interesting samples, commonly called \emph{most informative} samples, from a set of unlabelled data points, which are then passed to, e.g., a human annotator for labelling. The most informative samples are those, of which the model is most uncertain about. By actively choosing samples and asking for the correct labelling, the model is able to learn from few labelled data points, which is advantageous especially when annotated datasets are not available.
 
 Within the domain of political text analysis, many different tasks lack large amounts of annotated data.  It has been already shown in the past that these tasks can benefit from the active learning:  e.g., stance detection \citep{10.1145/3132169}, topic modeling \citep{romberg2022automated}, speech act classification \citep{10306154} or toxic comment classification \citep{Miller_Linder_Mebane_2020}. 

 
 In this work, we examine how LLM-generated synthetic data can be used in an active learning setting to specialise a stance detection agent on a specific question. 

\section{Method}


Our method uses the standard stance detection pipeline where the starting point is a pre-trained language model such as BERT \citep{devlin-etal-2019-bert} 
that is fine-tuned on an annotated stance detection dataset.
 We propose two ideas to improve the performance of the stance detection model:
our first idea is to use the synthetic data to augment existing data to fine-tune the stance detection model.
This alone is already a powerful tool to improve the performance of a stance detection model by grounding it to a specific question.
Our second idea uses synthetic data as an oracle for active learning, where the synthetic data act as an ensemble of experts for the QBC approach (query by committee).
We first introduce notation and preliminaries to explain our method. Then we describe how we generate synthetic data and how we use it for active learning.

\subsection{Preliminaries}
In stance detection, we usually have a set of labelled data $\mathcal{D} = \{(x^{(i)}, y^{(i)})\}_{i=1}^I$ where $x^{(i)}$ is a statement and $y^{(i)}$ is the stance of the statement,
with $y^{(i)} \in \{0, 1\}=\mathcal{Y}$. In stance detection, there is an additional question (sometimes also called issue or target) linked to each comment $x^{(i)}$.   The set of all questions is written as $\mathcal{Q}$.   We view the stance detection model as a binary classification function $f: \mathcal{Q} \times \mathcal{X} \rightarrow \mathcal{Y}$, which can be split into two parts, $f=h\circ g$:
the first part is a pre-trained language model $g: \mathcal{Q} \times \mathcal{X} \rightarrow \mathcal{E}$ such as BERT \citep{devlin-etal-2019-bert} or RoBERTa \citep{liu2019roberta}.  The second part is a classifier $h:\mathcal{E} \rightarrow \mathcal{Y}$ on top of the pre-trained model $g$.
The pre-trained language model $g$ outputs an embedding $e^{(i)}\in\mathcal{E}$ given a question $q^{(i)}$ and an input $x^{(i)}$. The classifier $h$ is usually a linear layer or a multi-layer perceptron, which takes the embedding $e^{(i)}$ from $g$ as input and outputs a prediction $\hat{y}^{(i)}$.
The stance detection model is \emph{fine-tuned} by minimizing the cross-entropy loss between the predicted labels $\hat{y^{(i)}} = f(t^{(i)}, x^{(i)})$ and the actual labels $y^{(i)}$.

\subsection{Generating Synthetic Data for Stance Detection}
To generate $M$ synthetic samples, we employ a quantized version of the Mistral-7B-Instruct-v0.1 model\footnote{\url{https://huggingface.co/turboderp/Mistral-7B-instruct-exl2}} to generate comments on a specific question $q$, using the following prompt: 

\noindent\fbox{\parbox{\linewidth}{
"\small\texttt{A user in a discussion forum is debating other users about the following question: [q] The person is (not) in favor about the topic in question. What would the person write? Write from the persons first person perspective.}".
}}
For our purpose we have two labels, so we generate $M/2$ for each label. We denote the question-specific synthetic dataset as:
\begin{equation}
\label{eq:dsynth}
\mathcal{D}_{\text{synth}} 
     = \big\{(x_{\text{synth}}^{(m)}, 1) \big\}_{m=1}^{M/2} \cup  \big\{(x_{\text{synth}}^{(m)}, 0) \big\}_{m=1+M/2}^{M}
\end{equation}
where half of the $M$ synthetic data samples have \emph{positive} labels, i.e., are comments in \emph{favor} for the posed question, while the other
half is \emph{against}. In the experiments we will use $\mathcal{D}_{\text{synth}}$ to augment the existing fine-tuning dataset $\mathcal{D}$ and in the next section for active learning.

\subsection{Synthetic Data-driven Query By Comittee}
\label{sec:active_learning}
In the typical active learning scenario, we have a set of unlabelled data $\mathcal{D}_{u} = \big\{x_{u}^{(n)}\big\}_{n=1}^N$, 
where the goal is to choose the most informative samples $\mathcal{D}_{\text{ch}} \subset \mathcal{D}_u$ to label. 
Our approach is similar to ``Query by Comittee'' (QBC) \citep{10.1145/130385.130417}, where an ensemble of models is trained on the labelled data $\mathcal{D} = \{(x^{(m)}, y^{(m)})\}_{m=1}^M$ and then chooses the subset $\mathcal{D}_\text{ch}$ of $\mathcal{D}_{u}$ for labelling.
Instead of an ensemble of experts, we directly use the labelled dataset $\mathcal{D}$: each {labelled data sample} is used as an \emph{oracle} for the unlabelled data by 
checking the similarity between the embeddings of the unlabelled data to the labelled data (see below).   Note that we use the synthetic data from the previous section for $\mathcal{D}$ and thus call our method \emph{Synthetic data-driven QBC} or short SQBC.  Our method performs four steps:

\paragraph{\textbf{(1) Generate the embeddings.}}  Given some embedding function $g:\mathcal{X}\rightarrow\mathcal{E}$, we generate embeddings for the unlabelled data,
\begin{equation}E_u = \big\{e_{u}^{(n)}\big\}_{n=1}^N = \big\{g(x_{u}^{(n)}) \big\}_{n=1}^N \end{equation} 
and for the labelled data
\begin{equation}
E = \big\{e^{(m)}\big\}_{m=1}^M = \big\{g(x^{(m)})\big\}_{m=1}^M.
\end{equation}
We use cosine similarities to define the nearest neighbours in the space of the embeddings.
E.g., for the $n$th unlabelled data point $x_u^{(n)}$ let $NN(n)$ be the set of indices of the $k$ nearest neighbours (wrt. to the embeddings) among the labelled data set $\mathcal{D}$.  In this paper, we choose $k=M/2$ since that is the size of one of the classes for a balanced dataset of size $M$ with $2$ classes (see Eq.~\eqref{eq:dsynth}).

\paragraph{\textbf{(2) Using the $M/2$ nearest neighbours as oracles to score the unlabelled data.}}
The score for each unlabelled data point counts the number of labels $y^{(m)}=1$ among the nearest neighbours, i.e., 
\begin{equation}
    s(n) = \sum_{m\in NN(n)}y^{(m)} \in [0, \ldots, M/2].
\end{equation}

\paragraph{\textbf{(3) Choosing the most informative samples.}}
The scores take values between 0 and $M/2$.  For 0, the nearest neighbours all have labels $y^{(m)}=0$, for value $M/2$, all have labels $y^{(m)}=1$.  The \emph{most informative} samples have a score in between.  A threshold $\kappa$ controls how many of those we choose,
\begin{equation}
    \mathcal{D}_{\text{ch}} := \Big\{x_{u}^{(n)} | \min_{n'} s(n') + \kappa < s(n) < \max_{n'} s(n') - \kappa \Big\}.
    \label{eq:kappa}
\end{equation}
We select the samples that have scores in the middle range. 
This is because samples with scores near the extremes can be confidently categorized and thus offer little information. 
Conversely, samples with scores that are neither distinctly positive nor negative are the most informative. 
The hyperparameter $\kappa$ is used to control the number of samples that are chosen. Higher values of $\kappa$ result in fewer samples to label manually,
thus reducing the labelling effort. We also consider the not chosen samples $D_{nch} := D_u \setminus D_{ch}$.

\paragraph{\textbf{(4) Assigning labels.}}
In the experiments, the chosen data points $\mathcal{D}_\text{ch}$ will get their true label.  In some variants of our method we also label the not chosen examples $\mathcal{D}_\text{nch}$ using the majority vote of the $k$ nearest neighbours,
\begin{equation}\label{eq:labels}
    y_u{(n)} = \frac{s(n)}{M/2}.
\end{equation}
These labels should be reasonable estimates, since they align strongly with the synthetic data.

\begin{figure*}[htbp!]
  \centering
    \begin{tikzpicture}
      \node[draw, thin, color=white, fill=white, rounded corners=0.0mm, inner sep=1mm] (img2){
        \includegraphics[width=0.99\textwidth]{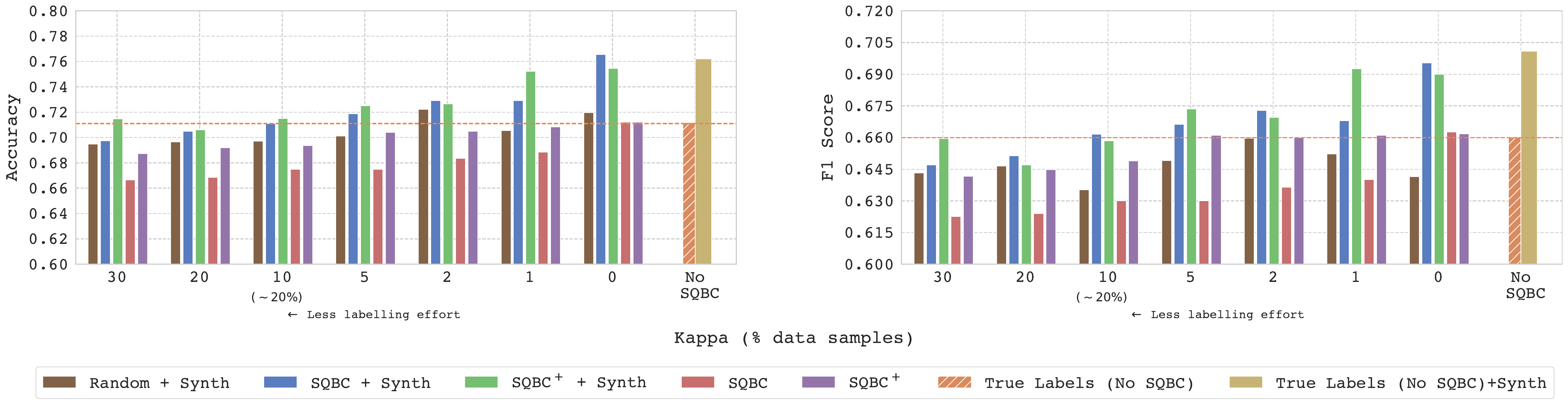}
      };
          \node[text width=7cm, text centered, above of=img1] at (-0.5, 1.7) {\fontfamily{pcr}\selectfont\textbf\scriptsize{Average Performance over 5 Questions}};	
        \end{tikzpicture}

\caption{\textbf{SQBC (Synthetic Data-driven Query By Comittee) can reduce labelling while improving stance detection performance:}
  Larger values of $\kappa$ correspond to fewer amount of samples chosen for manual labelling (see Eq. \eqref{eq:kappa}), hence reducing
  labelling effort. All bars above the dashed line 
  represent performance superior to the baseline. Using the chosen samples for manual labelling together with the remaining (not chosen) samples for fine-tuning 
  delivers the best overall performance. In some cases, labelling effort can be reduced substantially (by needing only 20\% of the data), 
  while still delivering better performance than the model with all true labels.  If a dataset is already labelled, then using synthetic data to augment 
  the dataset also improves the performance of the stance detection model.}
  \label{fig:results}
\end{figure*}

\section{Experiments}

\subsection{Dataset}
To evaluate our methods, we use the X-Stance dataset \citep{vamvas2020xstance}, a multilingual stance detection dataset, which contains a total of $67,000$ annotated comments on $194$ policy-related questions, answered by Swiss election candidates. X-stance includes comments in German ($48,600$), French ($17,200$) and Italian ($1,400$). The data has been extracted from smartvote\footnote{\url{https://www.smartvote.ch/}}, a Swiss voting advice platform. 
The comments are labelled either as being (i) in \emph{favor} or (ii) \emph{against} the posed question, which is a typical labelling scheme for stance detection datasets. Further details on the dataset can be seen in Appendix \ref{sec:dataset}.
We only use German comments, which are translated between English and German for the LLM. However, our approach can be applied to any other language where a good English translation is possible as well.


\subsection{Experimental Setup}
\label{sec:experimental_setup}
To measure the impact of each of variant of our active learning method SQBC on the performance of a stance detection agent, we propose different experimental setups. For all of the following setups, 
the starting model is the pre-trained BERT base\footnote{\url{https://huggingface.co/dbmdz/bert-base-german-uncased}} model,
fine-tuned on the X-Stance training dataset $\mathcal{D}_{\text{train}}$. We call this the \emph{base-model}. 
We then evaluate the performance of the base-model on the X-Stance test dataset $\mathcal{D}_{\text{test}}^{(q)}$ for different
questions $q$, which are not in the training set. We split each $\mathcal{D}_{\text{test}}^{(q)}$ into a $60/40$ train/test split. Consequently, we train all the following setups on the train split and then evaluate them on the test split:
\begin{itemize}[leftmargin=*]
    \item \textbf{True Labels w/o SQBC}: this variant assumes that all true labels for the train split of $\mathcal{D}_{\text{test}}^{(q)}$ are available and thus performs \emph{no active learning}.  This is the case $\mathcal{D}_\text{ch}=\mathcal{D}_u$.
    \item \textbf{True Labels w/o SQBC + Synth}: we additionally train on the synthetic data $\mathcal{D}_{\text{synth}}^{(q)}$. 
    \item \textbf{SQBC}: we apply our method on the train split of $\mathcal{D}_{\text{test}}^{(q)}$ to get $\mathcal{D}_{\text{ch}}^{(q)}$, for which we use the true labels. Then, we fine-tune our base-model on $\mathcal{D}_{\text{ch}}$ and evaluate it on the test split. 
    \item \textbf{SQBC + Synth}: same as SQBC, but we additionally use the synthetically generated data for the corresponding question to fine-tune the base-model.
    \item \textbf{SQBC$^+$}: as in SQBC, we apply active learning to get the chosen samples $\mathcal{D}_{\text{ch}}^{(q)}$, but we now additionally also use the non-chosen samples $\mathcal{D}_{\text{nch}}^{(q)}$ with their estimated label (see Eq.~\eqref{eq:labels}).
    \item \textbf{SQBC$^+$ + Synth}: same as SQBC$^+$, but we additionally use the synthetic data. 
    \item \textbf{Random + Synth}: as an ablation study, we replace SQBC with random selection, i.e., $\mathcal{D}_{\text{ch}}^{(q)}$ is randomly selected.
  \end{itemize}
A summary of all the variants can be found in Table \ref{tab:experimental_setup}. 
To study how the number of labelled samples impact the performance, we compare all method for varying values of $\kappa$, i.e., $\kappa\in \{0, 1, 2, 5, 10, 20, 30\}$.  The larger the value of $\kappa$, 
the fewer samples are chosen for $\mathcal{D}_{\text{ch}}$. The number of chosen samples determined by $\kappa$ is also used for the {Random + Synth} variant. As performance metric, we use the accuracy and the F1-score of
the stance detection model. We average the performance over 5 chosen questions in the X-Stance test set.

\begin{table}[h]
  \centering
  \scriptsize{
  \begin{tabular}{|l|c|c|c|}
  \hline
  \diagbox{Config}{Datasets} & \makecell{Manual labels \\ $D_{\text{ch}}$} & \makecell{Synth labels \\ $D_{\text{nch}}$} & \makecell{Synth Aug \\ $D_{\text{synth}}$} \\
  \hline
  True Labels &  &  &  \\
  \hline
  True Labels + Synth & & & \checkmark \\
  \hline
  SQBC  & \checkmark& &  \\
  \hline
  SQBC + Synth  &\checkmark&  & \checkmark\\
  \hline
  SQBC$^+$ & \checkmark&\checkmark &  \\
  \hline
  SQBC$^+$ + Synth & \checkmark & \checkmark & \checkmark  \\
  \hline
  Random + Synth & \checkmark (randomly selected) &  & \checkmark  \\
  \hline
  \end{tabular}
  }
  \caption{Synth: Synthetic Data, Aug: Augmentation. We compare different variants of active learning with synthetic data.}
  \label{tab:experimental_setup}
  \end{table}

\subsection{Results}
\paragraph{\textbf{Augmenting an existing fine-tuning dataset with synthetic data improves the overall performance.}}
To show the effectiveness of augmenting the fine-tuning dataset with synthetic data, 
we compared the performance of the stance detection model with true labels (striped orange bar) with and without synthetic data.
We see that augmenting the fine-tuning dataset with synthetic data (yellow bar) improves the performance of the stance detection model. 
In this sense, if there is a dataset available that is already labelled,
this data set can still be augmented with synthetic data to improve the performance of the stance detection model.

\paragraph{\textbf{SQBC reduces the labelling effort by leveraging synthetic data.}}
We compared the different variants of SQBC with and without synthetic data as described in Section \ref{sec:experimental_setup}. 
Larger values of $\kappa$ signify a smaller amount of samples chosen for manual labelling, hence reducing labelling effort. The corresponding
sample sizes for each value of $\kappa$ can also be seen in Figure $\ref{fig:results_extended}$. Furthermore, all bars above the dashed line 
represent performance superior to the baseline.  Active learning based on SQBC and also on SQBC$^+$ generally outperforms
the baseline. In some cases, we only need approximately $20\%$ of thedata while still delivering better performance as can be seen for $\kappa=10$. 
Using $\mathcal{D}_{\text{ch}}$ together with estimated labels for $\mathcal{D}_{\text{nch}}$  (green bar) for fine-tuning delivers the best overall performance.
As an ablation we showed that the samples selected by SQBC are more informative than randomly selected examples (brown bar).  Note that the random selection
strategy does not exceed the baseline in most cases. Finally, augmenting all variants of active learning with synthetic data further increases performance.


\section{Summary}
We presented two inherently linked approaches to improve stance detection models for online political discussions utilizing LLM-generated synthetic data:
first, we showed that augmenting the fine-tuning dataset with synthetic data related to the question of the stance detection improves the performance of the stance detection model.
Second, we showed that we can reduce the labelling effort by using synthetic data for active learning.  Furthermore, our active learning approach is superior to random selection of samples for manual labelling, and that augmenting all variants of active learning with synthetic data is essential for the best performance.
In the future, we will apply our ideas of using synthetic data for active learning to other tasks in natural language processing.



\bibliography{example_paper}
\bibliographystyle{icml2024}

\newpage
\appendix
\onecolumn

\section{Dataset}
\label{sec:dataset}
X-Stance is a multilingual stance detection dataset, including comments in German ($48,600$), French ($17,200$) and Italian ($1,400$) on political questions, answered by election candidates from Switzerland. The data has been extracted from smartvote\footnote{\url{https://www.smartvote.ch/}}, a Swiss voting advice platform. For the task of cross-topic stance detection the data is split into a training set, including questions on 10 political topics, and a test set with questions on two topics that have been held out, namely \emph{healthcare} and \emph{political system}. 


\section{Chosen questions}
We present the 5 chosen questions for our experiments in Table \ref{tab:questions}. We show the original questions in German and their corresponding English translations by the translation model. 
Furthermore, we also show the (60 / 40 ) train/test split for each question. 
\begin{table*}[h]
  \centering
  \begin{tabular}{|p{0.3\textwidth}|p{0.3\textwidth}|p{0.3\textwidth}|}
  \hline
  Question in German & Question in English & Data Splits 60/40 (num samples)\\ \hline
  Sollen sich die Versicherten stärker an den Gesundheitskosten beteiligen (z.B. Erhöhung der Mindestfranchise)  & Should insured persons contribute more to health costs (e.g. increase in the minimum deductible)? & \textbf{FAVOR-TRAIN:} 146 \textbf{AGAINST-TRAIN:} 154 \textbf{FAVOR-TEST:}87 \textbf{AGAINST-TEST:} 113 \textbf{TOTAL:} 500\\ \hline
  Befürworten Sie ein generelles Werbeverbot für Alkohol und Tabak?  & Do you support a general ban on advertising alcohol and tobacco? & \textbf{FAVOR-TRAIN:} 19 \textbf{AGAINST-TRAIN:} 44  \textbf{FAVOR-TEST:} 10 \textbf{AGAINST-TEST:} 33 \textbf{TOTAL:} 106 \\  \hline
  Soll eine Impfpflicht für Kinder gemäss dem schweizerischen Impfplan eingeführt werden?  & Should compulsory vaccination of children be introduced in accordance with the Swiss vaccination schedule? & \textbf{FAVOR-TRAIN:} 34 \textbf{AGAINST-TRAIN:} 83  \textbf{FAVOR-TEST:} 21 \textbf{AGAINST-TEST:} 58 \textbf{TOTAL:} 196\\ \hline
  Soll die Aufenthaltserlaubnis für Migrant/innen aus Nicht-EU/EFTA-Staaten schweizweit an die Erfüllung verbindlicher Integrationsvereinbarungen geknüpft werden?  & Should the residence permit for migrants from non-EU/EFTA countries be linked to the fulfilment of binding integration agreements throughout Switzerland? & \textbf{FAVOR-TRAIN:} 68 \textbf{AGAINST-TRAIN:} 40  \textbf{FAVOR-TEST:} 34 \textbf{AGAINST-TEST:} 39 \textbf{TOTAL:} 181 \\ \hline
  Soll der Bund erneuerbare Energien stärker fördern?  & Should the federal government promote renewable energy more? & \textbf{FAVOR-TRAIN:} 111 \textbf{AGAINST-TRAIN:} 50  \textbf{FAVOR-TEST:} 70 \textbf{AGAINST-TEST:} 38 \textbf{TOTAL:} 269 \\ \hline
  \end{tabular}
  \caption{Chosen questions for stance detection in German and their English translation with class splits.}
  \label{tab:questions}
\end{table*}

\section{Extended Results}
\begin{figure*}[htbp!]
	\centering
		\begin{tikzpicture}
			\node[draw, thin, color=white, fill=white, rounded corners=0.0mm, inner sep=1mm] (img2){
				\includegraphics[width=0.99\textwidth]{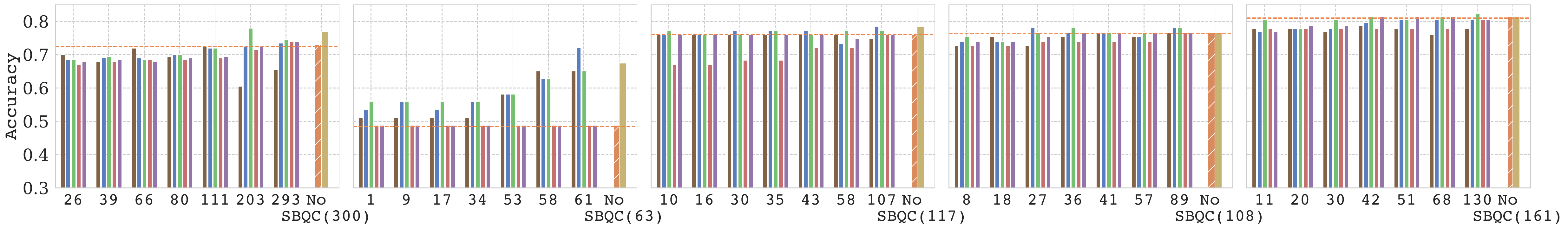}
			};
          \node[text width=2cm, text centered, above of=img1] at (-6.6, 0.5) {\fontfamily{pcr}\selectfont\textbf\scriptsize{Question 1}};	
          \node[text width=2cm, text centered, above of=img1] at (-3.2, 0.5) {\fontfamily{pcr}\selectfont\textbf\scriptsize{Question 2}};	
          \node[text width=2cm, text centered, above of=img1] at (0, 0.5) {\fontfamily{pcr}\selectfont\textbf\scriptsize{Question 3}};	
          \node[text width=2cm, text centered, above of=img1] at (3.5, 0.5) {\fontfamily{pcr}\selectfont\textbf\scriptsize{Question 4}};	
          \node[text width=2cm, text centered, above of=img1] at (6.8, 0.5) {\fontfamily{pcr}\selectfont\textbf\scriptsize{Question 5}};	

        \end{tikzpicture}
    
        \begin{tikzpicture}
			\node[draw, thin, color=white, fill=white, rounded corners=0.0mm, inner sep=1mm] (img2){
				\includegraphics[width=0.99\textwidth]{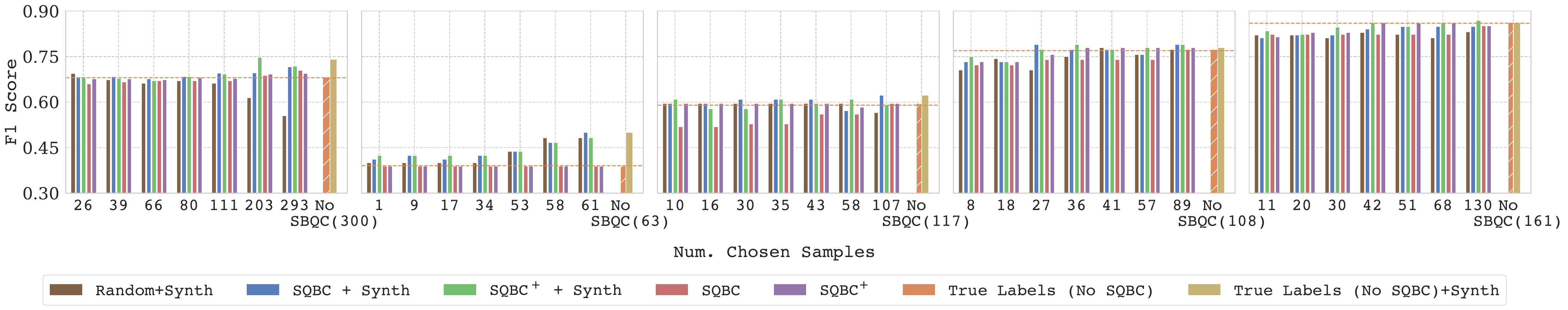}
			};
	    \end{tikzpicture}
    
\caption{\textbf{Extended results of Figure \ref{fig:results}:}
  We present the results for each of the 5 questions we chose from the X-Stance test set.
  The numbers on the x-axis correspond to the amount of samples chosen for manual labelling.
  These correspond to the values of $\kappa$ in Figure \ref{fig:results}.}
  \label{fig:results_extended}
\end{figure*}
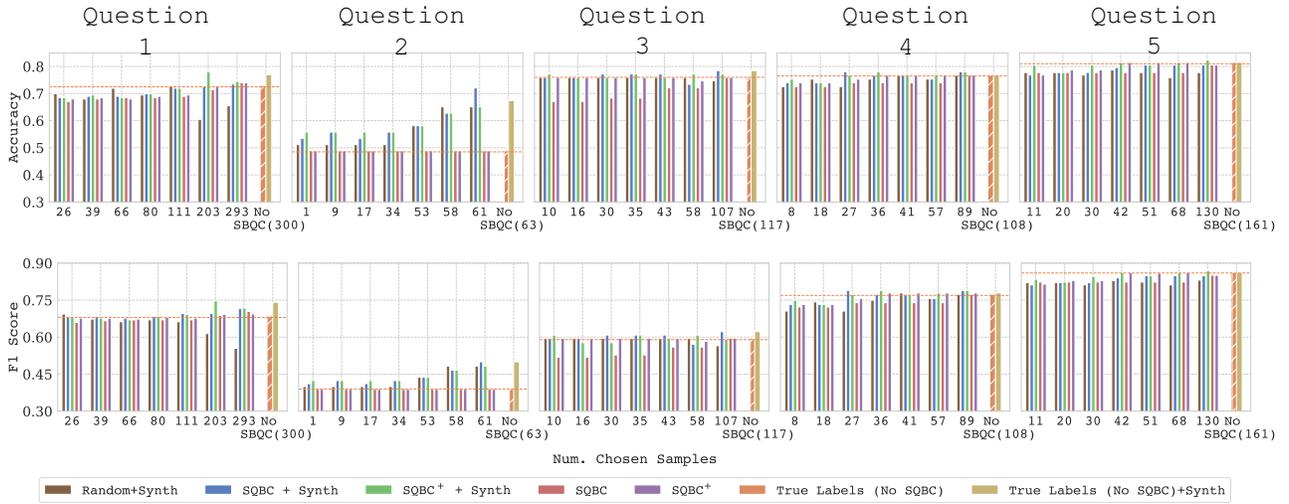

We show the results for each of the 5 questions we chose from the X-Stance test set in Figure \ref{fig:results_extended}. The numbers on the x-axis correspond to the amount of samples chosen for manual labelling. These correspond to the values of $\kappa$ in Figure \ref{fig:results}.
We see that question 5 does not benefit from augmenting the true labels. During training we observed that the model is able to better predict the stance for datasets where the split contains more favor than against labels. This can be seen for questions 4 and 5, 
where the augmentation with synthetic data of the true labels is not as effective. However, when using the synthetic data for active learning, the performance is improved for all questions, since in this scenario the manuall labelling set is smaller
and the class balance may change. Overall, we see that performance exceeding or similar to the baseline can be maintained with far fewer samples that need to be labelled. 

\end{document}